\documentclass[11pt,a4paper]{article}
\usepackage[hyperref]{emnlp-ijcnlp-2019}
\usepackage{times}
\usepackage{latexsym}
\usepackage{url}

\usepackage{mathptmx}
\usepackage{tikz-qtree}
\usepackage{pgfplots}
\usepackage{filecontents}
\usepackage{csvsimple}
\usepackage{amsfonts,amsmath}
\usepackage{url}
\usepackage{verbatim}
\usepackage{color}

\definecolor{lgray}{gray}{0.95}
\definecolor{lblue}{rgb}{0.90,0.90,1.00}
\definecolor{lyellow}{rgb}{1.00,1.00,0.70}

\usepackage{listings}
\newtheorem{ex}{Example}

\lstnewenvironment{code}
    {\lstset{}%
      \csname lst@SetFirstLabel\endcsname}
    {\csname lst@SaveFirstLabel\endcsname}
\newcommand{\BI}[0]{\begin{itemize}}
\newcommand{\EI}[0]{\end{itemize}}
\newcommand{\I}[0]{\item}
\newcommand{\BE}[0]{\begin{enumerate}}
\newcommand{\EE}[0]{\end{enumerate}}

\newcommand{\BX}[0]{\begin{ex}}
\newcommand{\EX}[0]{\end{ex}}
\newcommand{\BV}[0]{\begin{verbatim}}
\newcommand{\EV}[0]{\end{verbatim}}

\newcommand{\BF}[0]{\begin{filecontents*}{data.csv}}

\newcommand{\BQ}[0]{\color{blue}\begin{quote}}
\newcommand{\EQ}[0]{\end{quote}\color{black}}

\def \bscale1 {0.25}
\def \bscale {0.25}

\newcommand{\FIG}[4]{
\begin{figure}[htbp]
\centering
{\includegraphics[scale=#3]{figs/#4}}
\caption{#2}
\label{#1}
\end{figure}
}

\newcommand{\DFIG}[4]{
\begin{figure*}[htbp]
\centering
{\includegraphics[scale=#3]{figs/#4}}
\caption{#2}
\label{#1}
\end{figure*}
}

\aclfinalcopy 

\title{Dependency-based Text Graphs for Keyphrase and Summary Extraction with Applications to Interactive Content Retrieval}

\author{ Paul Tarau \\
  {\tt paul.tarau@unt.edu} \\\And
  Eduardo Blanco \\
  {\tt Eduardo.Blanco@unt.edu} \\
}

\begin{document}
\date{}
\maketitle

\begin{abstract}

We build a bridge between  neural network-based machine learning and graph-based natural language processing and introduce a unified approach to keyphrase, summary and relation extraction by aggregating dependency graphs from links provided by a deep-learning based dependency parser. 

We reorganize dependency graphs to focus on the most relevant content elements of a sentence, integrate sentence identifiers as graph nodes and after ranking the graph, we extract our keyphrases and summaries from its largest strongly-connected component. We take advantage of the implicit structural information that dependency links  bring to extract subject-verb-object, is-a and part-of relations. 

We put it all together into a proof-of-concept  dialog engine that specializes the text graph with respect to a query and reveals interactively the document's 
most relevant content elements. The open-source code of the integrated system is available at
\url{https://github.com/ptarau/DeepRank}.

{\bf Keywords}: {\em 
graph-based natural language processing,
dependency graphs,
keyphrase, summary and relation extraction,
query-driven salient sentence extraction,
logic-based dialog engine,
synergies between neural and symbolic processing.
}
\end{abstract}

\section{Introduction}\label{intro}


Recursive ranking algorithms applied to text graphs have enabled extraction of keyphrases, summaries and relations. Their popularity continues to increase due to the holistic view they shed on the interconnections between text units  that act as recommenders for the most relevant ones, as well as the comparative simplicity of the algorithms. 

At more that 2600 citations and a follow-up of some almost equally as highly cited papers like \cite{Erkan:2004} the TextRank algorithm \cite{EMNLP:TR,ijcnlp05} and its creative descendants have been used to extract summaries, keyphrases and relations from  several document types and social media interactions in a few dozen languages.

TextRank extracts keyphrases using a co-occurrence relation, controlled by the distance between word occurrences: two vertices are connected if their corresponding lexical units co-occur within a sliding window of 2 to 10 words.
Sentence similarity is computed as content overlap.
 For both keyphrases and summaries it uses a graph centrality algorithm like PageRank \cite{page98pagerank,brin98anatomy} to extract the  highest ranked text units. To accommodate the use of similarity relations, TextRank also introduces weights to the links that refine the original PageRank algorithm, but it needs elimination of stop words and reports best results when links are restricted to nouns and adjectives.
Independently, in \cite{Erkan:2004} several graph centrality measures are explored and \cite{radabook} offers a comprehensive overview on graph-based natural language processing and related graph algorithms.

Our  motivation for revisiting here the main assumptions of the ``TextRank-family of algorithms'' is to explore  synergies
 of graph-based natural language processing systems  with recent advancements in machine-learning driven models able to reveal deeper syntactic and semantic relations between text units. 
The main challenge in this quest is the right interface that exposes the machine-learned syntactic and semantic information to symbolic post-processing algorithms. 

This brings us to today's dependency parsers \cite{stan,AdolphsXLU11,choi:17a}.
Among them, the neurally-trained Stanford dependency parser \cite{stan} produces highly accurate dependency graphs and part of speech tagged vertices.
 Seen as edges in a text graph, they provide, by contrast to collocations in a sliding window,  ``distilled'' building blocks through which
 a graph-based natural language processing system can absorb
 higher level linguistic information.

Commitment to dependency relations, as they are 
associated to each sentence, 
entails also 
revisiting the
selection of the text units and links used to build text graphs,
as follows:
\BI
\I
Instead of using co-occurrence in a sliding window of a few words, we will use, for each sentence, dependency relations provided by the  dependency parser.
\I
Instead of deriving links from similarity relations between sentences based on bags of shared words, we will integrate sentence identifiers in the word-to-word graphs.
\EI

With help from the link labels and the part of speech tags associated to dependency types, we  extract highly-ranked facts and rules corresponding to  logical elements present in sentences that
we pass to logic programs that can infer new relations, beyond those that can be mined directly from the text.

We put our algorithms to a potentially practical use case: a dialog engine. Given a document, this engine answers queries as sets of salient sentences, by specializing our summarization algorithm to the context inferred from the query.

To summarize, the  most significant contributions of the research work covered by the paper are:
\BI
\I a unified algorithm for keyphrase, summary and relation extraction
\I state-of-the-art performance on  keyphrase and summary extraction, fast and scalable to large documents

\I a logic relation post-processor supporting realtime interactive queries 
about a document's content
\I integration of our algorithms into an open-source system with practical uses helping a reader of a scientific document to interactively familiarize herself with its content
\EI

The rest of the paper is organized as follows.
Section \ref{rel} overviews related work and background information.
Section \ref{kas} describes our unified keyphrase and summary extraction algorithm, with its results discussed and evaluated in 
section \ref{eval} .
Section \ref{dia} describes our open-source software
system's architecture and the interaction between the Python based document processor and the Prolog-based dialog engine.
Section \ref{conc} concludes the paper.

\section{Related work and research background}\label{rel}

We will focus our literature review on the key topics closely related to our goals, but refer the  reader to \cite{radabook} for a much wider spectrum of techniques and applications that have emerged in the quickly evolving field of graph-based natural language processing.

\subsubsection*{Dependency parsing}

The Stanford neural network based dependency parser  \cite{stan} is now part of the Stanford CoreNLP toolkit at \url{https://stanfordnlp.github.io/CoreNLP/} which also comes with  part of speech tagging, named entities recognition and co-reference resolution \cite{coreNLP}.
Its evolution toward the use of Universal Dependencies \cite{ud14} makes
tools relying on it potentially portable to
over {\bf 70}  languages  covered by the Universal Dependencies effort
\footnote{\url{https://universaldependencies.org/}}.

Of particular interest is the connection of dependency graphs  to semantic labeling and logic elements like predicate argument relations \cite{Choi:2011} which is present (together with features similar to the Stanford parser) in the NLP4J dependency parser implementation at \url{https://emorynlp.github.io/nlp4j/}. The mechanism of automatic conversion of constituency trees to dependency graphs described in \cite{choi:17a} allows the output of high-quality statistically trained phrase structure parsers to be reused for extraction of dependency links.

\subsubsection*{Leveraging knowledge resources}

The use of ranking algorithms in combination with WordNet synset links for word-sense disambiguation goes back as far as \cite{coling04:pr}, in fact a prequel to the TextRank paper \cite{EMNLP:TR}.

With the emergence of resources like Wikipedia, a much richer set of links and content elements has been used in connection with graph based natural language processing  
\cite{wikiRank,AdolphsXLU11,wikify}.

\subsubsection*{Keyphrase and summary extraction}

Text summarization techniques, including graph based ones are surveyed in \cite{NenkovaM12} and more recently in \cite{textSum}. Besides ranking, 
elements like coherence via similarity with previously chosen sentences
and avoidance of redundant rephrasings are shown to contribute to the overall quality of
the summaries. 

Beyond summaries obtained by aggregating important sentences extracted from a document, and possibly applying to them sentence compression techniques that remove redundant or less relevant words, new techniques are emerging for abstractive summarization \cite{abstractiveBing}. In the context of graph-based processing, one clearly benefits from as much syntactic and semantic information as possible, given also the need to synthesize new sentences subject to syntactic and semantic constraints.

\subsubsection*{Relation Extraction}

The relevance of dependency graphs for relation extraction has been identified in several papers, with \cite{AdolphsXLU11} pointing out to their role as a generic interface between parsers and relation extraction systems. 

In \cite{depRelPat} several  models grounded on syntactic patterns are identified (e.g., subject-verb-object) that can be mined out from dependency graphs.
Of particular interest for relation extraction facilitated by dependency graphs is the shortest path hypothesis that prefers relating entities like predicate arguments that are connected via a shortest paths in the graph \cite{Bunescu:2005}.

To facilitate their practical applications to biomedical texts, 
\cite{biorel} extends dependency graphs with focus on richer sets of semantic features including ``is-a'' and ``part-of'' relations and co-reference resolution.

\section{Our unified keyphrase and abstract extraction algorithm}\label{kas}

Unlike the approaches we have overviewed that develop special techniques for each of the tasks, we will present here a unified algorithm to obtain graph representations of documents, and show that this unified representation is suitable for keyphrase extraction, summarization and question answering.

We have organized our Python-based abstract and keyphrase extraction algorithm together with the Prolog-based dialog engine
into a  unified system.
Our implementation is available
\url{https://github.com/ptarau/DeepRank}.
We will overview first its Python-based document processing functionalities.

\subsection{Building the text graph}

We connect as a Python client to the CoreNLP server and use it to provide
our dependency links.

We  use the Python wrapper at \url{https://www.nltk.org/} for the Stanford CoreNLP toolkit \cite{coreNLP} and  launch the toolkit to run in server mode. 

We use unique sentence identifiers and unique lemmas as nodes of the text graph.

As  keyphrases are centered around nouns and good summary sentences are likely to talk about important concepts,  we will need to reverse some links in the dependency graph provided by the parser to prioritize nouns
and deprioritize verbs, especially auxiliary and modal ones. 
Thus we redirect the dependency edges toward nouns with subject and object roles, as shown for a simple short sentence  in {\bf Figure \ref{depgraph}}
as {\em ``about''} edges.
 
\DFIG{depgraph}{Dependency graph of a simple sentence with redirected  and  newly added arrows }{0.34}{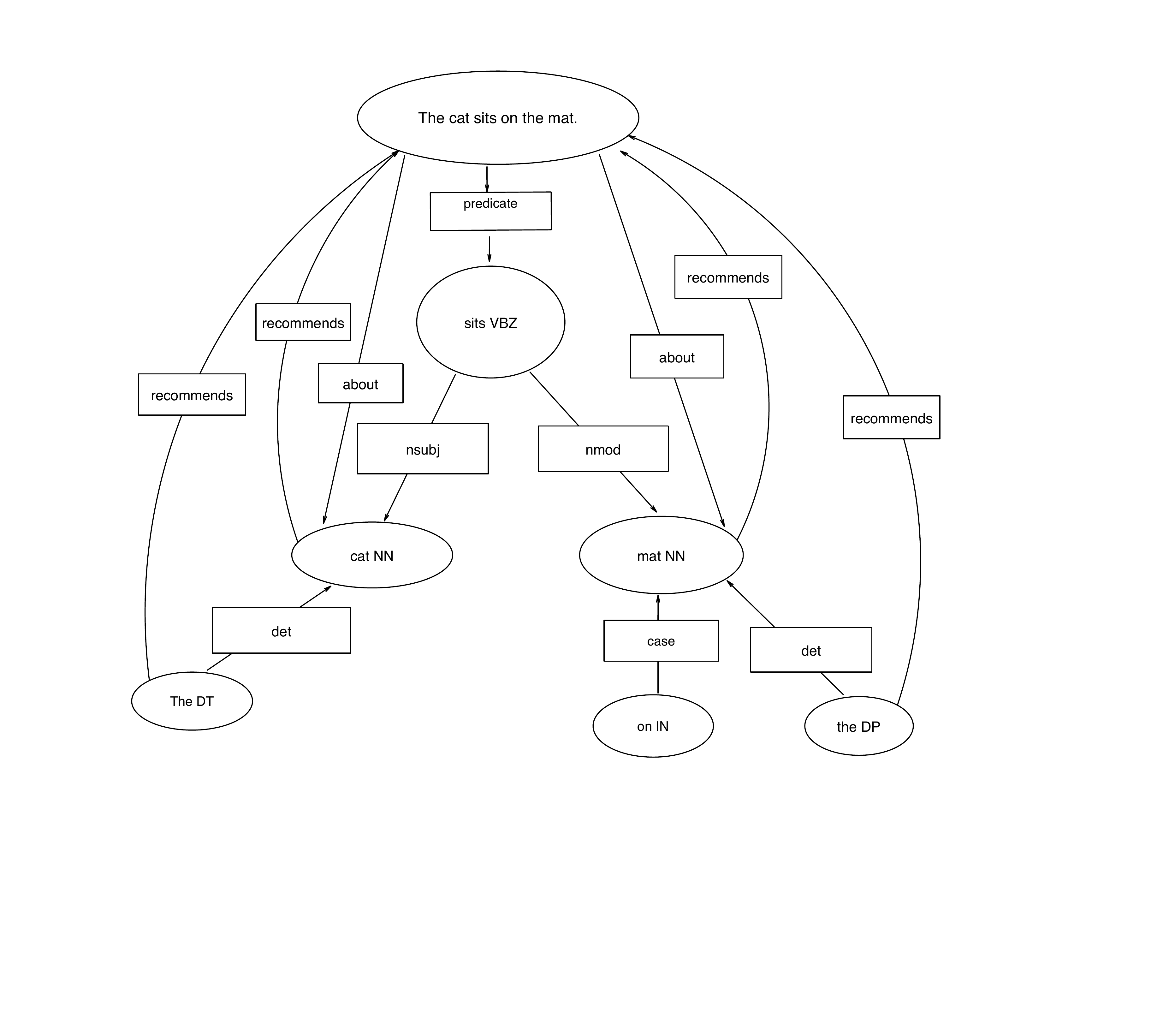}

We also create {\em ``recommend''} links  from words to the sentence identifiers and back from sentences to verbs with {\em predicate} roles
to indirectly ensure that sentences recommend and are 
recommended by their content. Specifically, we ensure that sentences 
recommend verbs with predicate function
from where their recommendation spreads to nouns relevant
as predicate arguments (e.g., having subject or object roles).

\subsection{Focusing the link graph on large strongly connected components}
We prioritize the largest strongly connected component of the text graph, aware that {\em emergence of giant components} \cite{giant} is likely to happen in a densely connected graph. The intuition behind this is that it is likely to help focus on the ``main story line'' of a text document. 

As link configurations  tend to favor very long sentences, a post-ranking normalization is applied for sentence ranking. We compute it as $new\_rank=rank/(1+2*sentence\_length)$.

 

\subsection{Pre- and post-ranking graph reductions}

The algorithm induces
 a form of automatic stopword filtering, 
due to the fact that our dependency link arrangement ensures that modifiers with lesser semantic value relinquish their rank by pointing to more significant lexical components.
This is a valid  alternative to explicit ``leaf trimming''  before ranking, which remains an option for reducing graph size for large texts or multi-document collections as well as helping with a more focussed relation extraction from the reduced graphs.
On the other hand, we provide a subgraph with trimmed low ranked nodes, after running a ranking algorithm, to
facilitate  human visual inspection of the resulting graphs 
(as it will be shown later in {\bf Figure \ref{constit}}). 

\subsection{From keywords to keyphrases}

We use the parser's compound phrase tags to  fuse along dependency links. Future work is planned to also use the Stanford  named entity recognizer to fuse named entities into single lexical units.
We   design our keyphrase synthesis algorithm to ensure that highly ranked  words will pull out their contexts from sentences, to make up meaningful keyphrases. As a heuristic, we mine for a context of 2-4 dependency linked words of a highly ranked noun, while ensuring that the context itself has a high-enough rank, as we compute a weighted average favoring the noun
over the elements of its context.

Besides extracting keywords and summaries, by using the PageRank implementation of the {\bf networkx} toolkit at \url{https://networkx.github.io/}, the system is also able to display
 a text graph filtered for its highest ranked vertices, using Python's 
 {\tt graphviz} library.

\section{Evaluation of our keyphrase and summary extraction algorithms}\label{eval}

Besides word-to-word links, our text graphs connect sentences as additional dependency graph nodes, resulting in a unified keyphrase and summary extraction framework. Our reliance on graphs provided by dependency parsers builds a bridge between deep neural network-based machine learning and graph-based natural language processing, 
often able to capture  implicit semantic information.

The algorithm works rather well on documents for which the underlying parser was trained like news articles and Web pages, but we wanted also to see how it performs  on a more complex older document, with a significantly different writing style and structure.

\subsection{An illustrative example}

We picked the {\em U.S. Constitution}
\footnote{\url{https://www.usconstitution.net/const.txt}},
 a fairly difficult text, with very long sentences.

From it, our algorithm extracted as summary, the sentence set \{1,2,3,5,6\}.
Its overlap with the popular baseline taking sentences from the beginning of a document is fairly accidental as we have not seen it 
on most of the other documents we have processed.
Besides the fact that each of those sentences is quite relevant, we explain this by the focus of the US Constitution on the legislative branch, that has overall the largest weight in terms of details in the document and by the choice of sentences belonging to the largest strongly connected component of the text graph.

The program has also extracted the following keyphrases { {\em 
['States', 'Washington', 'state', 'Congress', 'President', 'person', 'Amendment', 'law', 'power', 'House', 'year']}}, hinting  at the three branches of the U.S. government, and catching some of the key entities it discusses. The word-to-word graph connecting the highest ranked words is shown in {\bf Figure \ref{constit}}. 

As an interesting fact, coming from the use of a graph centrality algorithm like PageRank, our algorithm spots out the single dependency link that relates {\em Washington} to {\em president}, bringing him into the keyword set, by contrast to none of the other names mentioned in the document.

Overall, the graph
gives an idea on the effectiveness of the recommendation mechanism
generating directed edges from dependency links.

While this  hints to the fact that the proposed approach is working relatively well, it also suggests that  improvements are expected by integrating the label information (e.g., by adding higher weights to predicate-argument link candidates)  as well as the co-reference resolution and  named entity recognition made available by the parser.

\subsection{Quantitative Evaluation}

\FIG{constit}{Text graph of highest ranked words in the U.S. Constitution}{0.36}{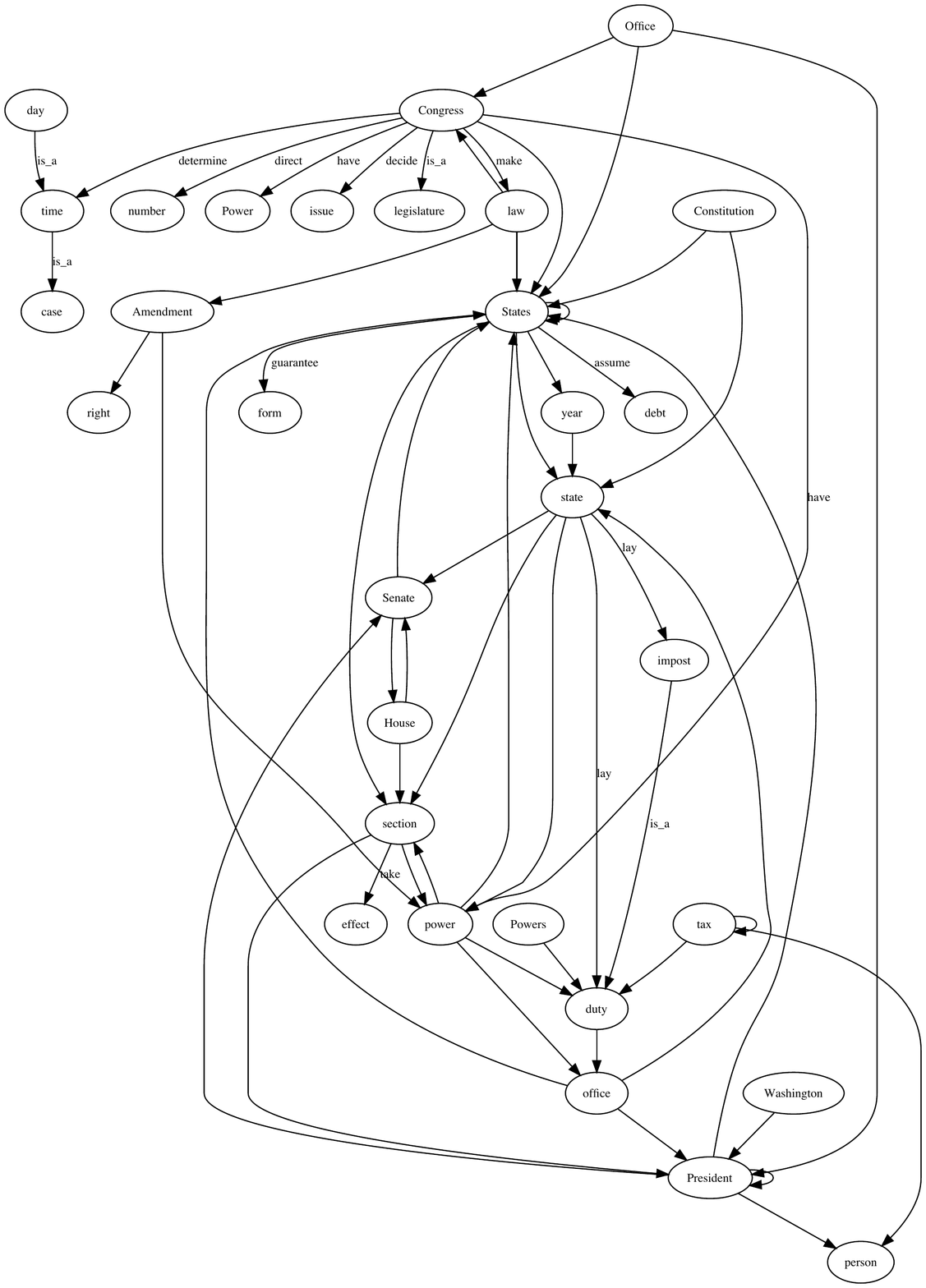}

\begin{figure}
\begin{center}
\begin{tabular}{||l||r|r|r||r||}
\cline{1-4}
\cline{1-4}
\multicolumn{1}{||l||}{{Abs. removed}} & 
  \multicolumn{1}{l|}{{ Precision}} &
  \multicolumn{1}{l|}{{ Recall}} &
  \multicolumn{1}{l|}{{ F-meas. }} \\
\cline{1-4} 
Keyphrases:10 & 0.25 & 0.25 & {\bf 0.24}  \\ \cline{1-4}
Sentences:9 & 0.31 & 0.43 & 0.35 \\ \cline{1-4}
Rouge-1:9 & 0.35 & 0.45 & {\bf 0.38}  \\ \cline{1-4}
Rouge-l:9 & 0.30 & 0.37 & 0.32  \\ \cline{1-4}
\end{tabular} \\
\medskip
\caption{Keyphrase (10) and abstract (9 sentences) extraction metrics, with abstract removed
\label{perf2}}
\end{center}
\end{figure}

\begin{figure}
\begin{center}
\begin{tabular}{||l||r|r|r||r||}
\cline{1-4}
\cline{1-4}
\multicolumn{1}{||l||}{{Full text}} & 
  \multicolumn{1}{l|}{{ Precision}} &
  \multicolumn{1}{l|}{{ Recall}} &
  \multicolumn{1}{l|}{{ F-meas. }} \\
\cline{1-4} 
Keyphrases:10 & 0.25 & 0.26 & {\bf 0.24}  \\ \cline{1-4}
Sentences:9 & 0.38 & 0.52 & 0.43 \\ \cline{1-4}
Rouge-1:9 & 0.40 & 0.54 & {\bf 0.44}  \\ \cline{1-4}
Rouge-l:9 & 0.37 & 0.48 & 0.40  \\ \cline{1-4}
\end{tabular} \\
\medskip
\caption{Keyphrase (10) and abstract (9 sentences) extraction metrics, full text
\label{perf3}}
\end{center}
\end{figure}

For an as accurate as possible evaluation of the keyphrase and summary extraction, we focus on testing against the {\em authors' own keyphrases and abstracts}, available with the {\em Krapivin document set} \cite{dataset08} repository, using  a Python implementation\footnote{https://github.com/Diego999/py-rouge} of
Rouge \cite{rouge}
for the summaries.

For evaluation of keyphrases, we have 
developed a simple tool using the {\tt nltk} implementation  of
{\em precision, recall and F-measure} with stemming applied to sets of words.

We verify it against computation by Rouge on summaries, on which it returns, more conservatively, slightly lower scores, thus giving us good confidence that it will not lift up the evaluation of the extracted keyphrases.

 {\bf Figure \ref{perf2}} shows results on the {\bf 2304} files of {\em computer science  papers} with author abstracts removed. {\bf Figure \ref{perf3}} uses the complete original document, which results in some improvement, especially on the summary extraction metrics.
Comparison is done against the authors' own abstracts and keyword sets,
which we consider more reliable than what third party human annotators could provide. We choose a large dataset of rather difficult scientific documents
as we want to test our algorithms on a realistic and potentially useful target for practical applications.
In both tables, rows {\tt 1-4} cover recall and F-measure computed by us for the extracted  keyphrases and sentences and then the Rouge-1 and Rouge-l metrics. Numbers after the colons indicate the size (e.g., 9,10) of the keyphrase and sentence set.

Our results for {\em summaries} on full documents are comparable with the
the state of the art\footnote{http://nlpprogress.com/english/summarization.html}.

Based on a recent review for keyphrase extraction
\cite{keys18} on the {\em Krapivin
document set}, our algorithm outperforms all others by a significant margin,
with values below {\bf 0.20} for  other graph-based systems.
Its performance 
is slightly below the best system in the set, {\bf CopyRNN}, 
which reports an F1-measure of {\bf 0.25} for a keyphrase set of size {\bf 10},
the same size as  the one we have run our tests.
Note also that our algorithm, including the input of the file and the call to the parser took only a few seconds on each of the documents in the collection. As a scalability test, processing a {\bf 250 pages book} took {\bf 28} seconds on a Mac Pro laptop.

\section{The relation processor and the dialog engine}\label{dia}

We have designed our system's summary and keyword extraction mechanism as a first step toward building a dialog engine, that, given a document, can sustain a meaningful conversation about it with an interested reader. 

As a use case, after hearing the summary and the keyphrases the document is about, a researcher or a student would ask specific questions that specialize the summary extraction to reveal the most pertinent sentences in the text related to the user's interest. One can see this as a ``mini search-engine'', specialized to a document and, with help of a document indexing layer, extensible to a multi-document collection.

\subsection{Generating  input for post-processing by logic programs}

Ideally, one would like to evaluate the quality of natural language understanding of an AI system by querying it not only about a set of relations explicitly extracted in the text, but also about relations inferred from the text. Moreover, one would like also to have the system justify the inferred relations in the form of a proof, or at least a sketch of the thought process a human would use for the same purpose.
The main challenge here is not only that theorem-proving logic is hard, (with first-order classical predicate calculus already Turing-complete), but also that modalities, beliefs, sentiments, hypothetical and contrafactual judgements often make the underlying knowledge structure intractable.

On the other hand, simple relations, stated or implied by text elements that can be mined or inferred from a ranked graph built from labeled dependency links, provide a limited but manageable approximation of the text's deeper logic structure, especially when aggregated with generalizations and similarities  provided by WordNet or the much richer Wikipedia knowledge graph. 

Once the document is processed, we generate,
besides the dependency links provided by the parser,
relations containing   essential facts we have gleaned 
from processing the document.

To keep the interface simple, we pass the following predicates in the form of Prolog-readable code, in one file per document. The predicates are:
\BI
\I keyword(WordPhrase). --  the extracted keyphrases
\I summary(SentenceId,SentenceWords). --  the extracted summary sentences sentence identifiers and list of words
\I dep(SentenceID, WordFrom, FromTag, Label, WordTo, ToTag). -- a component of a dependency link, with the first argument indicating the sentence they have been extracted
\I edge(SentenceID, FromLemma, FromTag, RelationLabel, ToLemma, ToTag). --  edge marked with sentence identifiers indicating  where it was extracted from, and the lemmas with their POS tags at the two ends of the edge

\I rank(LemmaOrSentenceId, Rank). -- the rank computed for each lemma
\I w2l(Word,Lemma,Tag). -- a map associating to each word a lemma, as found by the POS tagger
\I svo(Subject,Verb,Object). -- {\bf SVO} relations extracted from parser input or WordNet-based {\tt is\_a} and {\tt part\_of} labels in verb position 
\I sent(SentenceId,ListOfWords). -- the list of sentences in the document with a sentence identifier as first argument and list of words as second
\EI
They provide a relational view of a document in the form of a database that  will support  the inference mechanism built on top of it.

Currently we  add subject-verb-object facts extracted from the highest ranked dependency links, enhanced with ``is-a'' and ``part-of'' relations using WordNet via the {\tt nltk} toolkit, but we plan in the future to also generate relations from conditional statements  identified following dependency links and involving negations, modalities, conjuncts and disjuncts, to be represented as Prolog rules. We limit relations derived from the wrong synset, we constrain the two ends of an ``is-a'' or ``part-of'' edge to occur in the document. This provides a simple but effective of word-sense disambiguation heuristic.

The resulting logic program can then be processed  with Prolog semantics enhanced by using constraint solvers, abductive reasoners or Answer Set Programming systems \cite{asp}.

\subsection{The Prolog interface}

Currently we use as a logic processing tool the popular open source SWI-Prolog  system\footnote{\url{http://www.swi-prolog.org/}}
 \cite{swi} that
 can be called from, and can call Python programs using
 the {\tt pyswip} adaptor\footnote{https://github.com/yuce/pyswip}.
 After the adaptor creates the Prolog process and the content of the 
 digested document
 is transferred from Python, usually in a few seconds for typical 
 scientific paper sizes of 10-15 pages, query processing is realtime.

\subsection{The dialog engine}

Our Prolog-based dialog engine activates a conversational agent associated to a given document that answers queries as sets of salient sentences extracted from the text, via a specialization of our summarization algorithm to the context inferred from the query.

As part of an interactive  {\em read/listen, evaluate, print/say} loop,
we generate for each query sentence, a set of predicates that
are passed to the Prolog process, from where answers will come back
via the {\tt pyswip} interface.
The predicates have the same structure as the 
database 
representing the content of the complete document,
initially sent to Prolog.

\subsection{The query-answering mechanism}

Our query engine can be seen as a specialization of the summary extraction algorithm presented in Section \ref{kas}. Specifically, the engine returns the highest-ranked sentences based on relevance to the query and in the order in which they appear in the document.

After digesting the document the dialog will be about,
the dialog engine displays summary and keyphrases extracted from the document
\footnote{And also speaks them out if the {\tt quiet} flag is off.}.

We start with a simple query-expansion mechanism
via {\bf SVO} relations. They are derived by
finding, for lemmas in the query,
WordNet hypernyms, hyponyms, meronyms and holonyms,
as well as by directly extracting them from the query's dependency links.

Next, on the Prolog side, matching against the database
is currently implemented as  as a size constraint on the 
intersection of the expanded
query lemma set, built with highly ranked shared lemmas pointing to sentences
containing them.

We keep the dialog window relatively small (limited to the highest ranked 3 sentences in the answer). 
Relevance is ensured with help from the rankings computed for both
the  document content and the query. 

The following example queries the US Constitution document.

{\small
\begin{quote}
{\verb~>>>~ dialog\_about('examples/const',None)}

{\em
{\bf \verb~?--~ How can a President be removed from office?}

59 : In Case of the Removal of the President from Office , or of his Death , Resignation , or Inability to discharge the Powers and Duties of the said Office , the same shall devolve on the Vice President , and the Congress may by Law provide for the Case of Removal , Death , Resignation or Inability , both of the President and Vice President , declaring what Officer shall then act as President , and such Officer shall act accordingly , until the Disability be removed , or a President shall be elected . 

66 : Section 4 The President , Vice President and all civil Officers of the United States , shall be removed from Office on Impeachment for , and Conviction of , Treason , Bribery , or other high Crimes and Misdemeanors . 

190 : If the Congress , within twenty one days after receipt of the latter written declaration , or , if Congress is not in session , within twenty one days after Congress is required to assemble , determines by two thirds vote of both Houses that the President is unable to discharge the powers and duties of his office , the Vice President shall continue to discharge the same as Acting President ; otherwise , the President shall resume the powers and duties of his office .
}
\end{quote}
}
Note the relevance of the extracted sentences and resilience to semantic and syntactic variations.

The following example uses an ASCII version of
Einstein's 1920 book on relativity, retrieved from the Gutenberg
collection\footnote{
{\small \url{https://www.gutenberg.org/files/30155/30155-0.txt}}
}
and trimmed to the actual content of the book (250 pages in {\tt epub} form).

{\small
\begin{quote}
{\verb~>>>~ dialog\_about('examples/relativity',None)}

{\em
{\bf \verb~?--~ What happens to light in the presence of gravitational fields?}

611 : In the example of the transmission of light just dealt with , we have seen that the general theory of relativity enables us to derive theoretically the influence of a gravitational field on the course of natural processes , the laws of which are already known when a gravitational field is absent . 

764 : On the contrary , we arrived at the result that according to this latter theory the velocity of light must always depend on the co-ordinates when a gravitational field is present . 

765 : In connection with a specific illustration in Section XXIII , we found that the presence of a gravitational field invalidates the definition of the coordinates and the time , which led us to our objective in the special theory of relativity . 
}
\end{quote}
}

After the less that 30 seconds that it takes to digest the book, answers
are generated in less than a second for all queries that we have tried.
Given the availability of spoken dialog, a user quickly can iterate and refine
queries to extract the most relevant answer sentences of a document.

\section{Conclusions}\label{conc}

The key idea of the paper has evolved from our search for synergies between symbolic AI and emerging machine-learning based natural language processing tools. It is our belief that these are complementary and that by working together they will take significant forward steps in natural language understanding.

This has motivated us to experiment with a unified keyphrase extraction and summarization algorithm. The main novelty is to leverage dependency trees to generate text graphs instead of co-occurrences within a token window.

We have based our text graph on heterogeneous, but syntactically and semantically meaningful text units (words and sentences) resulting in a web of interleaved  links, mutually recommending each other's highly ranked instances.

Our  relation extraction algorithm, in combination with the Prolog interface, although still in work-in-progress stage, looks like a promising way to post-process the ranked syntactic and semantic information extracted from dependency graphs.

The dialog engine built on the Prolog side promises to evolve into a practical application by supporting spoken interaction with a conversational
agent that exposes salient content of the document driven by the user's interest.

Given the standardization brought by the use of {\em Universal Dependencies}, 
our techniques are  likely to be portable to a large number of languages.

\newpage

\bibliographystyle{../INCLUDES/acl_natbib}
\bibliography{../INCLUDES/nlp,../INCLUDES/nlp1,../INCLUDES/net,../INCLUDES/theory,tarau,../INCLUDES/proglang}

\end{document}